\documentclass{article}

\usepackage{microtype}
\usepackage{graphicx}
\usepackage{subcaption}
\usepackage{booktabs} 

\usepackage{hyperref}

\usepackage[preprint]{icml2026}

\usepackage{amsmath}
\usepackage{amssymb}
\usepackage{mathtools}
\usepackage{amsthm}

\usepackage[capitalize,noabbrev]{cleveref}

\theoremstyle{plain}

\theoremstyle{definition}

\theoremstyle{remark}

\usepackage[textsize=tiny]{todonotes}

\newcommand{\ourmodel}{PuYun-LDM~} %

\icmltitlerunning{\ourmodel for High-Resolution Ensemble Weather Forecasts}

\begin{document}

\twocolumn[
  \icmltitle{PuYun-LDM: A Latent Diffusion Model for 
  High-Resolution Ensemble Weather Forecasts}



  \icmlsetsymbol{equal}{*}

  \begin{icmlauthorlist}
    \icmlauthor{Lianjun Wu}{metac}
    \icmlauthor{Shengchen Zhu}{metac}
    \icmlauthor{Yuxuan Liu}{metac}
    \icmlauthor{Liuyu kai}{metac}
    \icmlauthor{Xiaoduan Feng}{metac}
    \icmlauthor{Duomin Wang}{metac}
    \icmlauthor{Wenshuo Liu}{metac}
    \icmlauthor{Jingxuan Zhang}{metac}
    \icmlauthor{Kelvin Li}{metac}
    \icmlauthor{Bin Wang}{metac}
  \end{icmlauthorlist}

  \icmlaffiliation{metac}{MetaCarbon} 
  \icmlcorrespondingauthor{Bin Wang}{21315079@zju.edu.cn}


  \vskip 0.3in
]



\printAffiliationsAndNotice{}  

\begin{abstract}
Latent diffusion models (LDMs) suffer from limited diffusability in high-resolution ($\leq 0.25^\circ$) ensemble weather forecasting, where diffusability characterizes how easily a latent data distribution can be modeled by a diffusion process. Unlike natural image fields, meteorological fields lack task-agnostic foundation models and explicit semantic structures, making VFM-based regularization inapplicable. 
Moreover, existing frequency-based approaches impose identical spectral regularization across channels under a homogeneity assumption, which leads to uneven regularization strength under the inter-variable spectral heterogeneity in multivariate meteorological data.
To address these challenges, we propose a 3D Masked AutoEncoder (3D-MAE) that encodes weather-state evolution features as an additional conditioning for the diffusion model, together with a Variable-Aware Masked Frequency Modeling (VA-MFM) strategy that adaptively selects thresholds based on the spectral energy distribution of each variable.
Together, we propose PuYun-LDM, which enhances latent diffusability and achieves superior performance to ENS at short lead times while remaining comparable to ENS at longer horizons. \ourmodel generates a 15-day global forecast with a 6-hour temporal resolution in five minutes on a single NVIDIA H200 GPU, while ensemble forecasts can be efficiently produced in parallel.

\end{abstract}

\begin{figure}[!t]
  \vskip 0.1in
  \begin{center}
    \centerline{\includegraphics[width=\columnwidth]{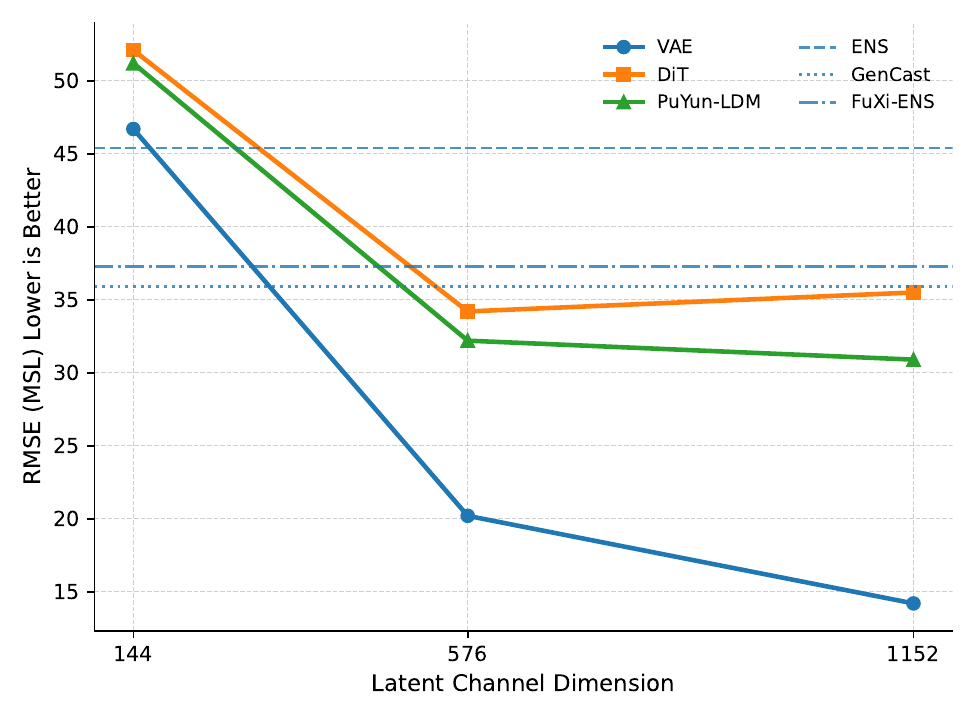}}
    \centerline{\includegraphics[width=\columnwidth]{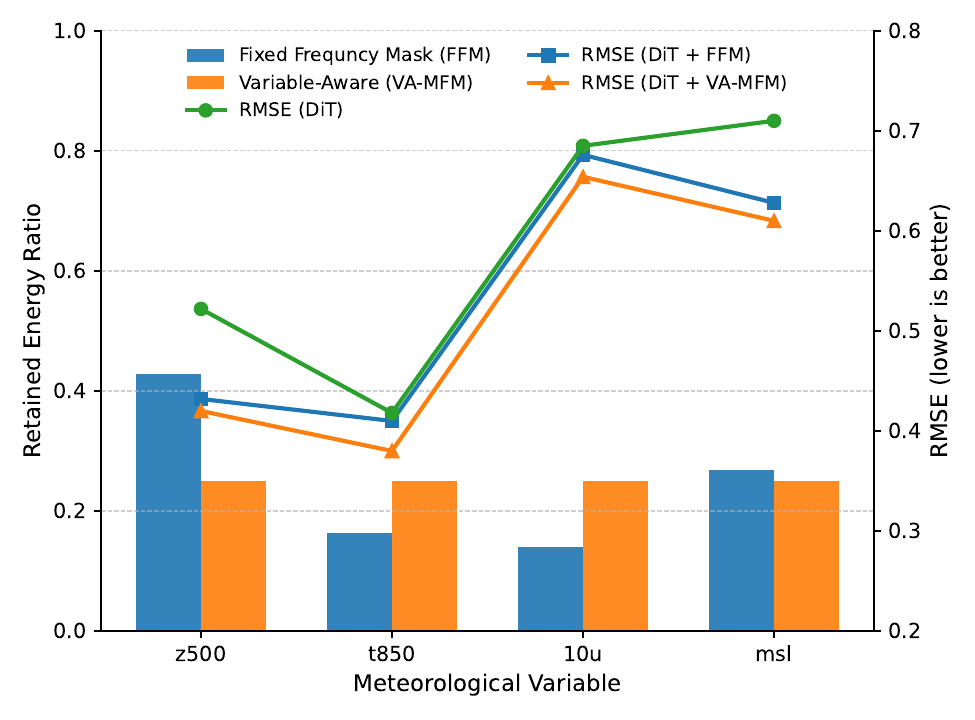}}
\caption{
\textbf{(Top)} Impact of latent dimensionality on MSL forecasting at the first lead time. 
Although VAE reconstruction error decreases with increasing latent dimension, DiT generation performance degrades at high dimensions, indicating reduced latent diffusability. 
\ourmodel effectively alleviates this issue.
\textbf{(Bottom)} Comparison of frequency-based regularization during VAE pretraining. 
FFM applies fixed low-pass thresholds of $0.25$ to latents and $0.05$ to inputs, whereas VA-MFM uses the same latent threshold but adopts adaptive, per-channel input thresholds.
FFM yield highly variable retained energy ratios across variables, resulting in imbalanced regularization.
RMSE values for Z500 and MSL are divided by $50$ for visualization clarity.
}
\label{fig:diffusability}
\vspace{-1.2cm}
  \end{center}

\end{figure}

\section{Introduction}
Medium-range high-resolution ensemble weather forecasting is critical for scientific and societal decision-making, as resolving fine-scale atmospheric structures while quantifying forecast uncertainty is essential for risk-sensitive applications in a chaotic system~\cite{palmer2002economic, leutbecher2008ensemble, alessandrini2011application, goodarzi2019decision}. Operational numerical weather prediction (NWP) systems therefore rely on ensemble forecasting—most notably the ECMWF Ensemble Prediction System (ENS)~\cite{molteni1996ecmwf}—to represent uncertainty via perturbed realizations. However, the prohibitive computational cost of running large ensembles at high spatial resolution severely limits ensemble size and degrades probabilistic skill for rare, high-impact events.

Recent advances in deep learning have enabled AI-based forecasting models to rival or surpass state-of-the-art NWP systems in deterministic prediction, motivating increasing interest in learning-based ensemble methods. Nevertheless, latent diffusion models (LDMs)~\cite{rombach2022high,karras_edm_2022,peebles2023scalable,wan2025wan}, despite their success in high-fidelity image generation, remain insufficiently developed for high-resolution ensemble weather forecasting. A central challenge lies in latent diffusability. As shown in Fig.~\ref{fig:diffusability}, training a high-capacity autoencoder on the meteorological data requires increasingly large latent feature dimensions to achieve acceptable reconstruction accuracy. While higher latent dimensionality improves reconstruction, it introduces a trade-off: generative performance of diffusion models saturates and even degrades~\cite{esser2024scaling,chen2025dc,kilian2024computational,yao2025reconstruction,skorokhodov2025improving}. This reconstruction–generation dilemma is also observed in meteorological settings (Fig.~\ref{fig:diffusability}), fundamentally limiting the applicability of LDMs to high-resolution ensemble forecasting.

In the vision domain, VA-VAE~\cite{yao2025reconstruction} attributes this degradation to an optimization difficulty arising from unconstrained high-dimensional latent spaces, and mitigates it by injecting the pretrained representations with high-level semantic information from foundation models such as MAE~\cite{he2022masked} and DINOv2~\cite{oquab2024dinov2learningrobustvisual}. However, meteorological data lack explicit semantic structures, and task-agnostic foundation models for atmospheric fields are currently unavailable. Motivated by the predictive nature of weather forecasting, we instead propose to constrain the latent space using temporal evolution features. Specifically, we introduce 3D-MAE, which masks the final weather state and reconstructs it from preceding states during pretraining. We hypothesize that this encourages the latent space to encode physically meaningful temporal evolution features, reducing the burden on the diffusion model to generate future states from unconstrained high-dimensional latents. As demonstrated in Fig.~\ref{fig:diffusability}, PuYun-LDM~\footnote{
PuYun means \emph{beautiful clouds} in Chinese, reflecting the traditional view that cloud patterns indicate future weather.} incorporates with 3D-MAE consistently improves diffusion-based forecasting accuracy, with larger gains at higher latent dimensions.

Beyond representation-alignment approaches, recent frequency-domain analyses~\cite{skorokhodov2025improving,chen2025dc,medi2025missing,lai2025toward,xiang2025denoising} reveal a fundamental cause of degraded latent diffusability: unstructured high-frequency components emerging in high-dimensional latent spaces conflict with the intrinsic low-frequency inductive bias of diffusion models. SE-VAE~\cite{skorokhodov2025improving} mitigates this issue via down-sampling-based regularization or a fixed frequncy mask strategy that suppresses high-frequency latent content. However, it implicitly assumes homogeneous spectral characteristics across channels. In multivariate meteorological fields, variables exhibit markedly different spectral profiles. As shown in the bottom panel of Fig.~\ref{fig:diffusability}, applying a fixed frequency threshold to inputs highly uneven fractions of spectral energy across different meteorological fields. We argue that this imbalance yields variable-dependent information content, resulting in unequal regularization strength and inconsistent performance gains. To address this limitation, we propose Variable-Aware Masked Frequency Modeling (VA-MFM), which adopts variable-specific frequency thresholds so that each field retains an equivalent fraction of spectral energy under the same regularization. By aligning the retained spectral energy ratio at the variable level, VA-MFM provides consistent regularization and more uniform improvements in downstream performance.

We summarize our contributions as follows:
\begin{itemize}
\item We introduce 3D-MAE, which encodes weather-state temporal evolution features as an additional conditioning for the conditional diffusion model and substantially improves latent diffusability.
\item  We propose VA-MFM, a variable-aware frequency-domain regularization that addresses the heterogeneous spectral characteristics of multivariate atmospheric fields.
\item  By integrating 3D-MAE and VA-MFM, we significantly improve the diffusability of high-dimensional latent spaces, leading to higher LDM-based forecasting accuracy in high-resolution ensemble weather prediction.
\end{itemize}

\begin{figure*}[t] 
  \vskip 0.2in
  \centering
  \includegraphics[width=0.9\textwidth]{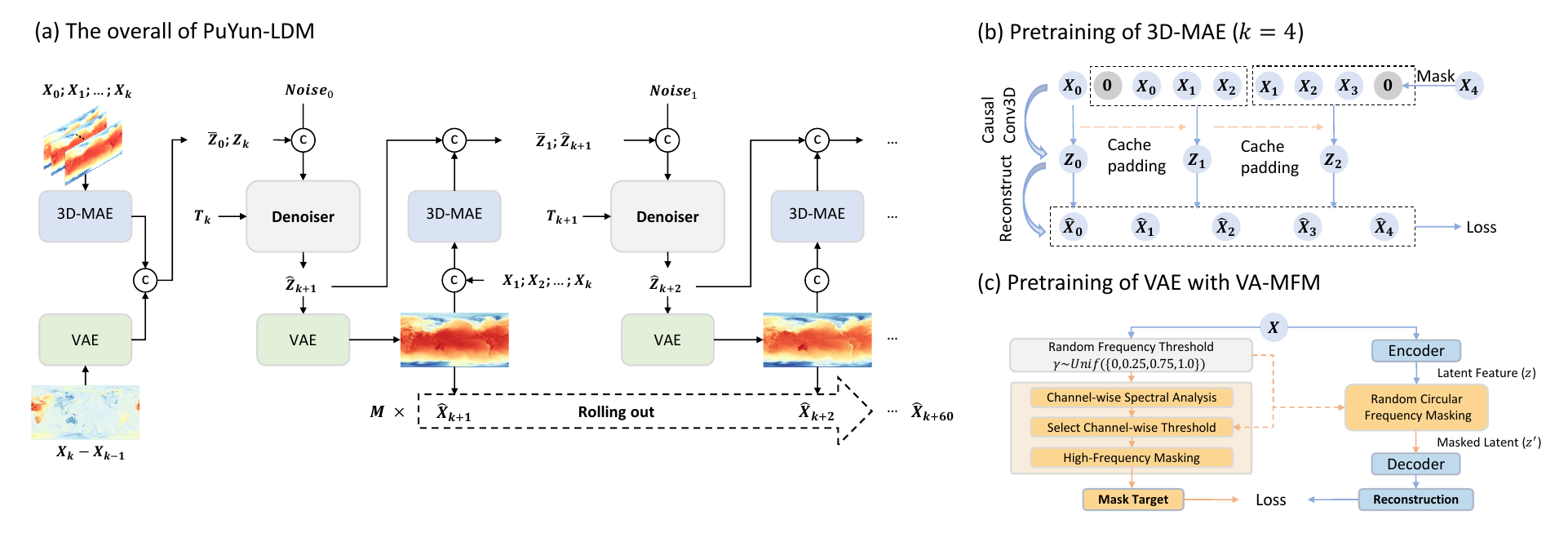}
\caption{Overview of the \ourmodel framework for ensemble weather forecasting.
(a) Overall architecture. Historical weather states are encoded by a VAE and a causal 3D-MAE, providing latent representations and temporal conditioning for an auto-regressive diffusion denoiser.
(b) Pretraining of 3D-MAE ($k=4$). The encoder uses causal 3D convolutions with caching to enforce temporal causality and extract temporal evolution features.
(c) Pretraining of VA-MFM. Channel-wise spectral analysis adaptively selects low-pass thresholds for the target, and high-frequency latent components are masked to suppress variable-specific artifacts and improve diffusability.}
  \label{fig:overall_pipeline}
  \vspace{-0.5cm}
\end{figure*}

\section{Related Work}

\paragraph{Medium-range Ensemble Forecasts Models.}

Operational ensemble forecasting systems such as ECMWF ENS~\cite{molteni1996ecmwf} remain the gold standard for uncertainty quantification but are computationally expensive. AI-based deterministic models, including FuXi~\cite{chen2023fuxi}, Pangu-Weather~\cite{bi2022pangu}, GraphCast~\cite{lam2022graphcast}, and PuYun~\cite{zhu2024puyunmediumrangeglobalweather}, enable efficient large-ensemble generation through stochastic perturbations but often suffer from unrealistic spread. Probabilistic models have therefore been introduced to explicitly model uncertainty. 
FuXi-ENS~\cite{zhong2025fuxi} adopts a VAE-based framework for ensemble weather forecasting, while GenCast~\cite{2023GenCast} formulates probabilistic prediction using conditional diffusion on an icosahedral mesh. LaDCast~\cite{zhuang2025ladcast} introduces latent diffusion into ensemble forecasting, and OmniCast~\cite{nguyen2025omnicast} explores a masked autoregressive formulation for probabilistic prediction. Despite the rapid emergence of diverse architectural approaches for high-resolution ensemble forecasting, latent diffusion model based frameworks remain relatively underexplored in this domain, particularly at global high resolutions.

\paragraph{Representation Alignment.}
To address the low diffusability of high-dimensional latent spaces, REPA-style~\cite{yu2024representation} methods have been proposed, which improve latent quality by introducing external encoders to accelerate diffusion convergence and enhance generative performance. VA-VAE~\cite{yao2025reconstruction} leverages vision foundation models to regularize latent spaces, yielding substantial gains in image generation. DDT~\cite{zheng2025diffusion} decouples diffusion transformer models incorporate a condition encoder to align the representation from VFMs and a velocity decoder to decode velocity. RAE~\cite{wang2025ddt} directly uses frozen pretrained representations from VFMs as the encoder with a lightweight decoder to reconstruct input images. However, in the meteorological domain, explicit semantic structures are absent, and task-agnostic foundation models analogous to DINOv2~\cite{oquab2024dinov2learningrobustvisual} are currently unavailable. These limitations fundamentally hinder the applicability of representation-alignment-based remedies to LDM-based ensemble weather forecasting.

\paragraph{Spectral-Based Methods}

SE-VAE~\cite{skorokhodov2025improving} analyzes the frequency characteristics (via DCT~\cite{ahmed2006discrete}) of inputs, reconstructions, and latents, identifying anomalous high-frequency components in the latent space as a key cause of diffusability degradation. To alleviate this issue, SE-VAE introduces a down-sampling-based regularization that enforces consistency between downsampled latents and downsampled inputs. Similarly, AFLDM~\cite{zhou2025alias} enforces shift equivariance in both the autoencoder and diffusion model, while EQ-VAE~\cite{kouzelis2025eq} introduces scale- and shift-equivariance regularization for autoencoders. FreqWarm~\cite{lai2025toward} highlights the dominant role of high-frequency components in reconstruction. Overall, these spectral-based methods implicitly assume homogeneous spectral characteristics across channels. However, in multivariate meteorological data, different variables exhibit distinct spectral profiles, causing uniform frequency constraints to affect variables unevenly and resulting in suboptimal diffusability improvements.

\section{\ourmodel}
\subsection{Preliminaries and Background}
Ensemble weather forecasting is commonly formulated as learning the conditional distribution of future weather states $X_{0:T}$ given past observations $O_{\le 0}$,
\begin{equation}
p(X_{0:T} \mid O_{\le 0}) .
\end{equation}
Under a first-order Markov assumption, this distribution factorizes as
\begin{equation}
p(X_{0:T} \mid O_{\le 0})
= p(X_0 \mid O_{\le 0}) \prod_{t=1}^{T} p(X_t \mid X_{t-1}),
\end{equation}
where $p(X_0 \mid O_{\le 0})$ is obtained via conventional NWP-based data assimilation~\cite{kalnay2003atmospheric,shin2016local,robert2017localizing}. We use ERA5~\cite{hersbach2020era5} reanalysis for initialization and evaluation. Our objective is to model the transition distribution $p(X_t \mid X_{t-1})$ using a conditional latent diffusion model~\cite{karras_edm_2022}. In practice, we observe that diffusion models exhibit substantially lower forecasting accuracy than VAE reconstructions, and this gap widens as the dimension of the latent space increases. 

To address these issues, we introduce two complementary strategies. First, 3D-MAE is designed to predict the next state from the past $k\!+\!1$ frames, thereby encoding temporal evolution features into the latent space and providing them as conditioning signals for the conditional diffusion model. This design effectively relaxes the first-order Markov assumption,
\begin{equation}
p(X_t \mid X_{t-1})
\;\approx\;
p\!\left(X_t \mid X_{t-1}, \ldots, X_{t-k-1}\right),
\end{equation}
as detailed in Sec.~\ref{sec:3dmae}. Second, VA-MFM regularizes the latent space by adaptively selecting variable-wise low-pass thresholds for the inputs and reconstructing them from correspondingly low-pass-filtered latents, thereby suppressing spurious high-frequency artifacts and ensuring balanced regularization strength across variables, as detailed in Sec.~\ref{sec:vamfm}.

The overall architecture of \ourmodel is illustrated in Fig.~\ref{fig:overall_pipeline}~(a). Our model produces 15-day forecasts at 6-hour intervals ($T=60$). Given an initial sequence $(X_0, \ldots, X_k)$ from ERA5, future weather states are generated autoregressively as
\begin{equation}
X_t \sim p\!\left(X_t \mid X_{t-1}, \ldots, X_{t-k-1}\right),
\end{equation}
for $t = k+1, \ldots, k+60$. This formulation naturally supports parallel sampling, enabling efficient generation of $M$ ensemble forecast members.

\subsection{Compression Model}
Our compression backbone is based on the Deep Convolutional Autoencoder (DC-AE)~\cite{chen_dcae_2025}, which is designed to support extremely high spatial compression ratios while maintaining reconstruction fidelity.
DC-AE combines hierarchical convolutional down-sampling with residual auto-encoding shortcuts, effectively mitigating the optimization and information bottlenecks that arise when encoding high-resolution fields.
This makes DC-AE particularly suitable for high-resolution meteorological data (e.g., $0.25^\circ$ or finer), where aggressive compression is required for scalable modeling.
Both our VAE and 3D-MAE adopt the DC-AE architecture and share its core encoder–decoder design.

\subsection{3D Masked AutoEncoder}
\label{sec:3dmae}
Inspired by advances in natural image generation~\cite{yu2024representation,yao2025reconstruction, zheng2025diffusion}, where incorporating pretrained representations that contain high-level semantic information has been shown to both accelerate diffusion convergence and improve diffusability, we propose to explicitly encode temporal evolution features into the latent conditioning.
To this end, we introduce a 3D Masked AutoEncoder (3D-MAE), which leverages temporal masking and reconstruction to force the model to learn causally structured spatiotemporal representations, rather than frame-wise spatial features.

Building on DC-AE, we replace all 2D convolutions with 3D convolutions and incorporate the caching mechanism of Wan-VAE~\cite{wan2025wan} to realize causal 3D convolutions. 
Formally, given $k{+}1$ consecutive weather states $(X_0, \ldots, X_k)$,
we apply temporal zero padding with three prepended frames, resulting in
$\hat{\mathbf{X}} = (\mathbf{0}, \mathbf{0}, \mathbf{0}, X_0, \ldots, X_k)$. Let $\mathrm{Conv3D}^{(l)}(\cdot)$ denote the $l$-th 3D convolutional layer,
and $[\cdot,\cdot]$ denote concatenation along the temporal dimension. 
For subsequent stages $s=1,2\ldots,1+\frac{k}{2}$, the model processes a temporal window
$[2s-2:2s+1]$ and incorporates cached features from the previous stage:
\begin{equation}
\left\{
\begin{aligned}
&\mathbf{h}_{s}^{(1)} =
\mathrm{Conv3D}^{(1)}\!\Big([\mathbf{\hat{X}}_{2s-2:2s+1}]\Big), \quad l=1, \\
&\mathbf{h}_{s}^{(l+1)} =
\mathrm{Conv3D}^{(l+1)}\!\Big([\mathbf{c}_{s-1}^{(l)},\, \mathbf{h}_{s}^{(l)}]\Big),
\quad l>1 ,
\end{aligned}
\right.
\end{equation}

where $\mathbf{c}_{s}^{(l)} = \mathbf{h}_{s}^{(l)}$ denotes the cached output at layer $l$ and stage $s$, and $\mathbf{c}_{0}^{(l)} = \mathbf{0}$. By default, we apply a temporal stride of $2$ in an intermediate 3D convolutional layer.

The pretraining procedure of 3D-MAE is illustrated in Fig.~\ref{fig:overall_pipeline}(b), where we show the forward process with $k=4$.
Given an input sequence of five consecutive weather states $\{X_0, X_1, X_2, X_3, X_4\}$, the model is trained in a temporally causal manner. In the final stage, the model processes the sequence $[X_1, X_2, X_3, X_4]$, where $X_4$ is replaced by an all-zero tensor $\mathbf{0}$.
The model is then required to reconstruct $X_4$ solely from the temporal features accumulated from the preceding states.
This masking strategy explicitly forces 3D-MAE to encode features related to temporal evolution, rather than relying on instantaneous spatial information, thereby producing more informative temporal latent representations.

Overall, the encoder realizes a causal mapping
\begin{equation}
\bar{z}_{t+k} = \mathcal{E}_{\text{3D}}(\tilde{X}_{t:t+k}),
\end{equation}
where $\tilde{X}_{t:t+k}$ denotes a causally padded input sequence in which the last frame is masked with zeros. The model is optimized using a masked reconstruction objective that reconstructs all $k{+}1$ frames in the input window:
\begin{equation}
\mathcal{L}_{\mathrm{3D\text{-}MAE}}
=
\mathbb{E}_{t}
\left[
\left\|
\mathcal{D}\!\left(\mathcal{E}_{3D}\!~(\tilde{X}_{t:t+k})\right)
-
X_{t:t+k}
\right\|_2^2
\right],
\end{equation}
where the causal masking in $\tilde{X}_{t:t+k}$ ensures that the reconstruction of future frames relies solely on past information. Note that we do not apply KL regularization in 3D-MAE.

\subsection{Variable-Aware Masked Frequency Modeling}
\label{sec:vamfm}

To address the issue of uneven regularization strength across variables in existing approaches, we introduce Variable-Aware Masked Frequency Modeling (VA-MFM), a frequency-domain regularization strategy designed for multivariate meteorological data.
VA-MFM explicitly aligns the low-frequency content between input fields and their latent representations while accounting for variable-specific spectral characteristics. Specifically, a masking ratio
\begin{equation}
\gamma \sim \mathcal{U}\bigl(\{0.25,\,0.5,\,0.75,\,1.0\}\bigr)
\end{equation}
is selected uniformly during VAE training.
When $\gamma < 1.0$, VA-MFM is applied; otherwise, the training reduces to standard VAE optimization.

For each meteorological variable $v$ (i.e., each input channel), let $X^{(v)} \in \mathbb{R}^{H \times W}$ denote the spatial field.
We compute its 2D Fourier transform
\begin{equation}
\mathcal{F}^{(v)} = \mathcal{F}\!\left(X^{(v)}\right),
\end{equation}
and obtain the radial amplitude spectrum $A^{(v)}(r)$ by averaging Fourier magnitudes over circular frequency shells.
The cumulative spectral energy is defined as
\begin{equation}
E^{(v)}(r) = \frac{\sum_{r' \le r} A^{(v)}(r')}{\sum_{r'} A^{(v)}(r')}.
\end{equation}
Given the selected ratio $\gamma$, we determine a variable-specific cutoff frequency
\begin{equation}
r^{(v)}_\gamma = \min \left\{ r \; \big| \; E^{(v)}(r) \ge \gamma \right\},
\end{equation}
which preserves a fixed proportion of spectral energy for variable $v$.

All frequency components above $r^{(v)}_\gamma$ are masked out, yielding a low-pass filtered target
\begin{equation}
\tilde{X}^{(v)} = \mathcal{F}^{-1}\!\left(
\mathcal{F}^{(v)} \odot M\bigl(r<r^{(v)}_\gamma\bigr)
\right),
\end{equation}
where $\odot$ is element-wise multiplication, and $M(r<r')$ is a circle mask to filter all frequency components above $r'$.

Symmetrically, the latent representation produced by the encoder $\mathcal{E}$ is subjected to the low-pass masking:
\begin{equation}
\tilde{\mathbf{z}} = \mathcal{F}^{-1}\!\left(
\mathcal{F}(\mathcal{E}(X)) \odot M\bigl(r<\gamma)
\right).
\end{equation}
Because all variables are highly coupled in the latent space, we apply variable-aware thresholds only on the reconstruction (image) side. The decoder is then trained to reconstruct $\tilde{X}$ from $\tilde{\mathbf{z}}$.
The loss of VAE is
\begin{equation}
\mathcal{L}_{\text{VAE}}
=
\underbrace{
\sum_v 
\left\|
\mathcal{D}(\tilde{\mathbf{z}}) - \tilde{X}^{(v)}
\right\|_2^2
}_{\text{reconstruction loss}}
+
\underbrace{
\beta \, \text{KL}\Big(q_\phi(\tilde{\mathbf{z}} \mid X) \;\|\; p(\tilde{\mathbf{z}})\Big)
}_{\text{KL regularization}},
\end{equation}  where $\beta$ is loss weight for KL regularization.


\subsection{Conditional Diffusion Model}
Conditioned on the latent of historical weather states, we use a diffusion model to generate the latent feature of the next-step weather state by denoising corrupted residual latents. Given a clean residual latent
$\mathbf{z}_{t+1}=\mathcal{E}(X_{t+1}-X_t)$ and a Gaussian noise
$\boldsymbol{\epsilon} \sim \mathcal{N}(\mathbf{0}, \mathbf{I})$:
\begin{equation}
\mathbf{z}_{t+1}' = \mathbf{z}_{t+1} + \sigma \boldsymbol{\epsilon},
\end{equation}
where $\sigma \sim p(\sigma)$, and $p(\sigma)$ is chosen as a log-normal distribution~\cite{karras_edm_2022}. The denoising network is trained to recover $\mathbf{z}_{t+1}$ from its noisy counterpart $\mathbf{z}_{t+1}'$.

\begin{figure*}[t] 
  \vskip 0.2in
  \centering
  \includegraphics[width=0.9\textwidth]{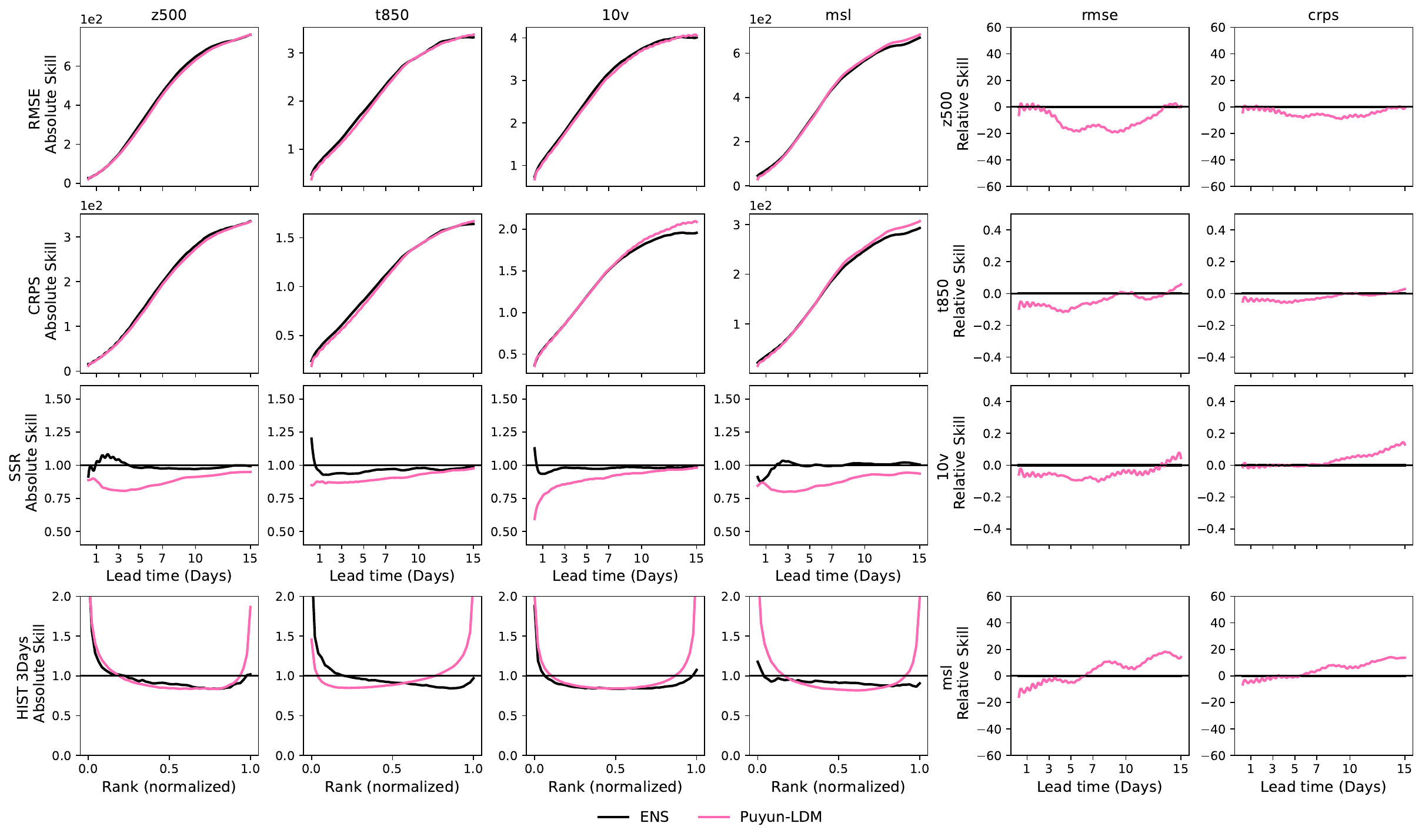}
  \caption{RMSE, CRPS, SSR, and Rank histograms of model comparison for z500, t850, 10v, and msl. For a fair comparison among models, we evaluate ENS against its corresponding analysis HRES-fc0 and \ourmodel against ERA5.}
  \label{fig:main_results}
  \vspace{-0.5cm} 
\end{figure*}

\paragraph{Conditioning.}
The input of the denoiser is formed by concatenating the noisy latent $\mathbf{z}_{t+1}'$ with a conditioning tensor $[\,\bar{z}_{\,t+1},\, z_t \,]$,
where $\bar{z}_{\,t+1}$ denotes the latent representation produced by the 3D-MAE encoder.
The concatenated tensor is reshaped into a 3D spatiotemporal form $(C,T,H,W)$, where
$T = 3 + \frac{k}{2}$ under a temporal stride of $2$.

\paragraph{Architecture.}
A $1{\times}1{\times}1$ 3D convolution first projects the concatenated input into a unified latent embedding space.
The network then applies 56 depth of DiT~\cite{peebles2023dit} blocks with 1152 hidden size, each consisting of RMS normalization, multi-head self-attention with fixed 3D rotary positional embeddings~\cite{su2024roformer}, and a SwiGLU FFN~\cite{shazeer2020glu}, connected through residual pathways.
Adaptive Layer Normalization, conditioned on timestep embeddings, modulates both attention and feed-forward submodules.
Finally, an RMSNorm~\cite{zhang2019root} followed by a linear projection maps the Transformer features back to the latent space, producing the predicted clean residual latent.

\paragraph{Objective.}
We train the denoiser by minimizing a weighted mean squared error between the predicted and ground-truth residual latents. The preconditioning functions and noise-dependent weighting terms are consistent with the original EDM~\cite{karras_edm_2022}.

\section{Experiments}
\paragraph{Data.}
The ERA5 reanalysis dataset~\cite{hersbach2020era5} is used in this study. 
ERA5 data are temporally subsampled to 6-hour intervals at the native $0.25^\circ$ resolution, using initial conditions at 00, 06, 12, and 18 UTC. 
The training set spans 1979--2017, with 2018 reserved for validation and 2019 for testing.
Model inputs include upper-air variables at 13 pressure levels (50, 100, 150, 200, 250, 300, 400, 500, 600, 700, 850, 925, and 1000\,hPa): geopotential (Z), specific humidity (Q), temperature (T), and horizontal wind components (U, V), together with surface variables including mean sea level pressure (MSL), 2-m temperature (T2M), and 10-m wind components (U10, V10). In total, 69 meteorological variables are used.

\paragraph{ENS and HRES-fc0.}
ECMWF provides two complementary forecasting products: the High-Resolution Forecast (HRES)~\cite{ecmwf_hres2019}, a single deterministic forecast, and the Ensemble Prediction System (ENS), which quantifies forecast uncertainty via perturbed initial conditions.
HRES-fc0~\cite{ecmwf_hres_fc0} denotes the 0-hour HRES forecast and serves as a dynamically consistent initial atmospheric state.
Operational ENS forecasts are generated by perturbing and evolving HRES-fc0.

\paragraph{Evaluation.}
Following WeatherBench2~\cite{rasp2024weatherbench}, we evaluate deterministic and probabilistic skill using the Root Mean Square Error (RMSE) and the Continuous Ranked Probability Score (CRPS).
RMSE measures the accuracy of the ensemble mean, while CRPS assesses the full predictive distribution, capturing both calibration and sharpness.
Metrics are computed grid-pointwise and globally averaged for each variable and pressure level. Probabilistic skill is assessed using the spread--skill ratio (SSR) and rank histograms. An SSR close to $1$ indicates well-calibrated ensemble spread.
Rank histograms assess ensemble reliability by measuring how often the verifying observation falls within different ranks of the ensemble forecast. A nearly flat histogram indicates that the ensemble spread is consistent with forecast uncertainty, whereas U-shaped or inverted U-shaped distributions reflect under- or over-dispersed ensembles, respectively.

\paragraph{Implementation Details.}
We train the VAE on standardized residual fields $\Delta \mathbf{X}=\mathbf{X}_t-\mathbf{X}_{t-1}$ to capture 6-hour atmospheric dynamics, using latitude-based area weighting and pressure-level-dependent loss weights~\cite{lam2022graphcast}. Optimization is performed with AdamW~\cite{loshchilov2017decoupled} ($\beta_1=0.9$, $\beta_2=0.95$), a constant learning rate of $5\times10^{-5}$, and $\lambda_{\mathrm{KL}}=1\times10^{-5}$ for $70$k iterations.

The 3D-MAE encoder is trained with a staged curriculum. It is first pretrained on single-frame low-resolution data ($180\times360$) for $30$k iterations (batch size $128$, LR $1\times10^{-4}$), followed by low-resolution sequences of five frames with the final frame masked for $15$k iterations (batch size $64$, LR $5\times10^{-5}$). Training then shifts to single-frame high-resolution data ($720\times1440$) for $30$k iterations (batch size $16$, LR $5\times10^{-5}$), and is finally fine-tuned on five-frame high-resolution sequences with the final frame masked for $15$k iterations (batch size $8$, LR $1\times10^{-5}$).

The diffusion model is trained for $100$k iterations using AdamW with a global batch size of $32$ and a constant learning rate of $1\times10^{-3}$.
At inference, we use $25$ denoising steps, with all other settings consistent with standard EDM. All three models are trained on eight NVIDIA H200 GPUs; the total training time is approximately 2.5 days for 3D-MAE, 2 days for the VAE, and 2 days for the diffusion model, respectively. Unless otherwise specified, both the VAE and 3D-MAE use a 1152-dimensional latent space and a 32× spatial downsampling.

\begin{table}[t]
  \caption{
  Ablation study on our proposed 3D-MAE and VA-MFM. Display the MSL results at the first lead time. FFM refers to Fixed Frequency Mask. The first row corresponds to a DiT model without using additional encoder outputs as conditioning inputs.
  }
  \label{tab:ablation_diffusability}
  \begin{center}
    \resizebox{\columnwidth}{!}{
    \begin{small}
      \begin{sc}
        \begin{tabular}{cccccc}
          \toprule
          2D-AE & 3D-MAE & FFM & VA-MFM & RMSE $\downarrow$ & SSR $\rightarrow 1$ \\
          \midrule
            &        &        &        & 43.2 & 0.806 \\
           $\surd$ &        &        &        & 37.7 & 0.823 \\
                  & $\surd$ &        &        & 35.5 & 0.876 \\
                  & $\surd$ & $\surd$ &        &  31.4  &  0.812  \\
                & $\surd$ &        & $\surd$ & \textbf{30.5} & 0.801 \\
          \bottomrule
        \end{tabular}
      \end{sc}
    \end{small}
    }
  \end{center}
  \vskip -0.1in
\end{table}

\begin{table}[t]
  \caption{Comparison of the performance between different spectral-based strategies across different meteorological variables. SE refers to Scale Equivariance. FFM refers to Fixed Frequency Mask. Here, we use DiT with 3D-MAE as the Baseline. RMSE of Q700 is multiplied by 1000.}
  \label{tab:spectral-based_strategies}
  \begin{center}
  \resizebox{\columnwidth}{!}{
    \begin{small}
      \begin{sc}
        \begin{tabular}{lccccccc}
          \toprule
          strategies & Z500 & T850 & U850 & Q700 & 2t & 10v & msl \\
          \midrule
          Baseline &    26.1    &   0.418    &    0.800    &   0.371   &    0.654      &   0.703     &    35.5    \\
          SE &   22.4    &    0.402    &   0.787     &    0.366    &    0.642    &  0.694      &     33.2   \\
          FFM &    21.6    &    0.410    &   0.779     &    0.362    &    0.657    &  0.700      &     31.4   \\
          VA-MFM(ours) &    \textbf{21.0}    &   \textbf{0.380}     &   \textbf{ 0.721 }   &   \textbf{0.344 }    &    \textbf{0.639}    &   \textbf{0.677}     &     \textbf{30.5}   \\
          \bottomrule
        \end{tabular}
      \end{sc}
    \end{small}}
  \end{center}
  \vskip -0.1in
\end{table}

\paragraph{\ourmodel for Medium-range Ensemble Weather Forecasting.}
Fig.~\ref{fig:main_results} reports 15-day autoregressive forecasts with a 6-hour interval, evaluated on two upper-air variables (Z500, T850) and two surface variables (10V and MSL). For each variable, we report RMSE, CRPS, SSR, and rank histograms across the forecast horizon. We calculated these metrics using the results from 48 ensembles for all models. 
Following standard practice in prior work~\cite{2023GenCast,zhong2025fuxi,nguyen2025omnicast}, we evaluate PuYun-LDM initialized from ERA5 and compare it with ENS initialized from HRES-fc0.
It is important to clarify that this comparison serves as a benchmark for assessing the predictive potential of the models, rather than a head-to-head evaluation under fully operational, real-time conditions. To ensure relative fairness, all metrics are computed with respect to each model’s corresponding analysis fields. 
Fig.~\ref{fig:main_results} show that, by capturing temporal evolution features through 3D-MAE and mitigating latent diffusability degradation with VA-MFM, PuYun-LDM achieves lower RMSE and CRPS than ENS at short lead times while remaining comparable to ENS at longer horizons. 

Note, all SSR results are reported using deterministic sampling. Following the stochastic sampler proposed in EDM~\cite{karras_edm_2022}, we can similarly introduce stochasticity into PuYun-LDM at inference time by injecting random noise into the deterministic ODE solver, trading a small increase in RMSE for improved SSR. Specifically, we adopt the EDM implementation with $\text{S\_churn}=2.5$, $\text{S\_min}=0.75$, $\text{S\_max}=68$, and $\text{S\_noise}=1.1$.
With stochastic sampling enabled, \ourmodel achieves improved probabilistic reliability on the first-step MSL prediction, increasing SSR from $0.801$ to $0.902$, while incurring only a marginal RMSE increase from $30.5$ to $30.9$.

\begin{figure}[!t]
  \vskip 0.1in
  
  \begin{center}
    \centerline{\includegraphics[width=\columnwidth]{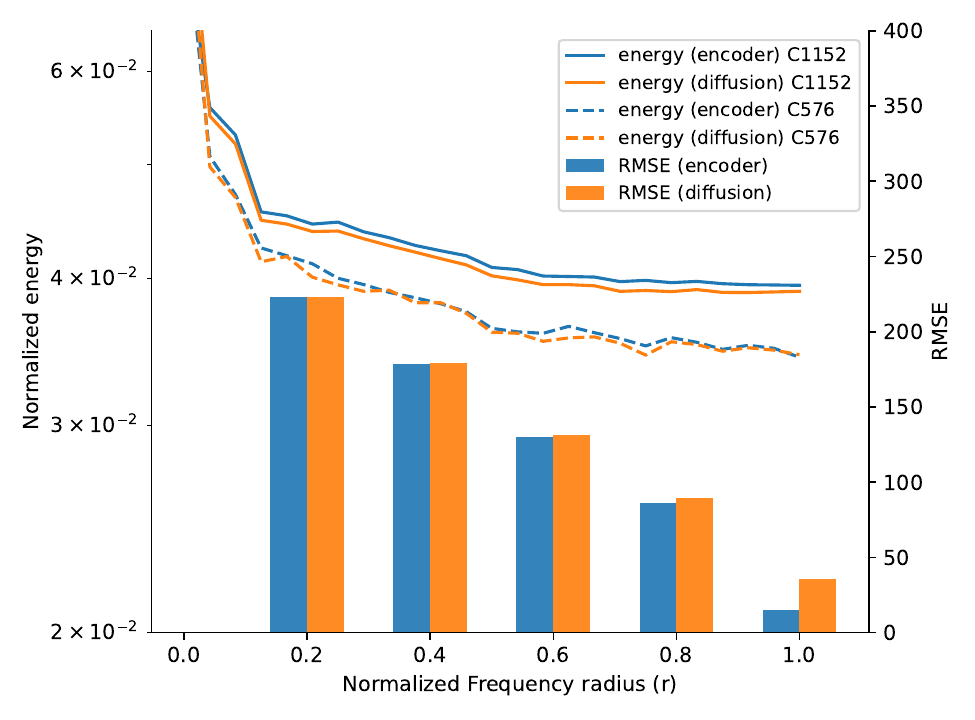}}
\caption{
Comparison of normalized spectral energy of encoder- and diffusion-generated latents across different frequency bands under varying latent dimensionalities, and the effect on RMSE of progressively masking high-frequency components.
}
\vspace{-0.55cm}
\label{energy_rmse}
  \end{center}
  
\end{figure}

\paragraph{Effect of 3D-MAE Conditioning.}
We investigate the effect of temporal latent conditioning through 3D-MAE. Tab.~\ref{tab:ablation_diffusability} compares DiT models equipped with 2D-AE and 3D-MAE conditioning, using identical latent dimensionality and conditioning channel budgets. Replacing 2D-AE with 3D-MAE consistently improves performance, reducing RMSE to $35.5$ and increasing SSR, which indicates better ensemble calibration. These results demonstrate that explicitly encoding temporal evolution in the latent space provides more informative conditioning for diffusion-based forecasting than frame-wise spatial representations alone. Moreover, comparing the first and second rows in Tab.~\ref{tab:ablation_diffusability}, we observe that the diffusion model can benefit from additional encoder-based conditioning. We attribute this to the fact that the VAE is trained on residual fields and therefore primarily captures temporal variations, whereas the additional encoder encodes largely static and complementary information.

\begin{figure*}[t] 
  \vskip 0.2in
  \centering
  \includegraphics[width=0.9\textwidth]{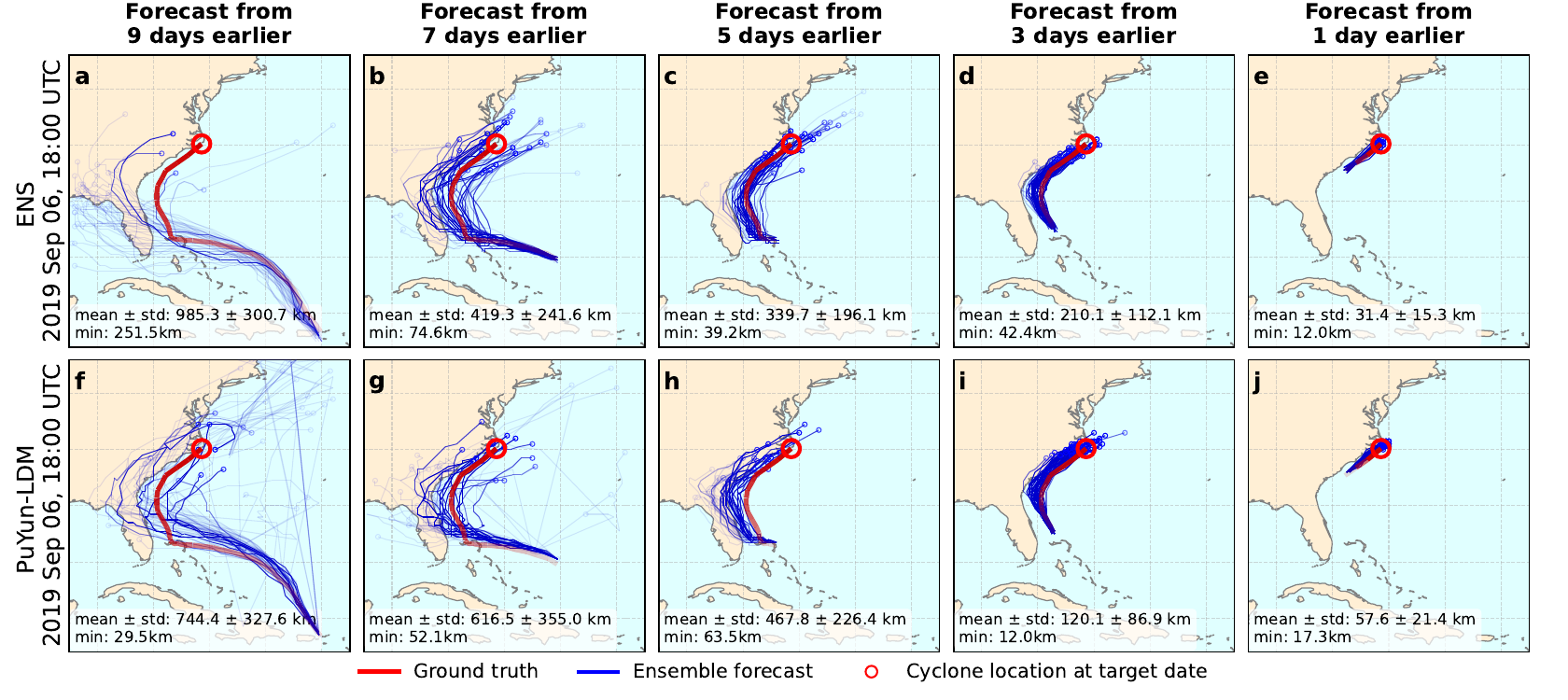}
  \caption{
Visualization of Hurricane Dorian trajectories at 18:00 UTC on September 6, 2019.
The blue curves show predicted track from \ourmodel and ENS initialized 1--9 days in advance. The mean, variance and minimum of the ensemble landfall error with respect to the observed landfall are shown in the lower-left corner of each panel. Predicted trajectories whose landfall locations deviate by more than 200 km from the observed landfall are rendered with reduced opacity for clarity.
}
  \label{fig:Dorian}
\end{figure*}

\paragraph{Effect of Variable-Aware Masked Frequency Modeling.}
Next, we analyze latent diffusability from a spectral perspective.
As shown in Fig.~\ref{energy_rmse}, the spectral energy of the diffusion latent closely matches that of the encoder latent in the low frequencies~($r<0.2$), while exhibiting systematically lower energy at high frequencies~($r \geq 0.2$). 
This pattern provides clear evidence of the \emph{intrinsic low-frequency bias} of diffusion models~\cite{falck2025fourier}. Fig.~\ref{energy_rmse} further illustrates the impact of progressively masking high-frequency components on RMSE for both latent representations.  Applying identical high-frequency suppression to the encoder latent leads to a much faster degradation compared to the diffusion latent. 
This discrepancy is most pronounced in the very high-frequency regime ($r > 0.8$), where the diffusion latent, already lacking extreme high-frequency information, is less affected by the frequency masking. Moreover, as the latent dimensionality increases, the relative contribution of extreme high-frequency components also increases. 
Since diffusion models struggle to learn these extreme high-frequency modes, increasing the latent dimensionality ultimately leads to degraded diffusion performance. Tab.~\ref{tab:spectral-based_strategies} compares three regularization strategies. SE introduces scale-equivariant regularization by reconstructing downsampled inputs from downsampled latents~\cite{skorokhodov2025improving}. FFM applies fixed-threshold low-pass filtering to both latents and inputs during VAE training. Specifically, we randomly remove latent frequencies above ${0.25, 0.5, 0.75, 1.0}$ while reconstructing inputs filtered above ${0.05, 0.10, 0.20, 1.0}$. Notably, the filtering thresholds for input reconstruction are not optimal. Adjusting the thresholds for the inputs can be seen as a step toward the adaptive frequency modulation implemented in VA-MFM. Both methods yield inconsistent improvements across variables due to spectral heterogeneity. In contrast, the proposed VA-MFM uses variable-adaptive frequency thresholds and achieves consistent positive improvements across all considered variables. 

\paragraph{Visualization of Tropical Cyclone Track.}
To evaluate the model's performance under extreme conditions, we visualized the tropical cyclone track of Hurricane Dorian, which stands as one of the most powerful and representative cyclones in the Atlantic basin. We apply the cyclone tracking algorithm, TempestExtremes v2.1~\cite{Ullrich2021_TempestExtremes}, using identical hyperparameters for both models. Tropical cyclone trajectories from the IBTrACS dataset~\cite{Knapp2010_IBTrACS} are used as ground truth. The predicted tracks and the ground truth are visualized in Fig.~\ref{fig:Dorian}.
In addition, we quantify track prediction errors by computing the mean, variance, and minimum of the positional deviation between the predicted landfall locations and the observed landfall location. The results show that \ourmodel produces more accurate landfall location predictions than ENS when forecasting from 3 and 9 days earlier, while achieving comparable performance to ENS at other lead times.

\section{Conclusion}
In this work, we propose PuYun-LDM, a latent diffusion framework for medium-range global weather forecasting, designed to improve latent diffusability under high-dimensional latent spaces. First, we introduce 3D-MAE, which provides temporally causal and information-efficient conditioning by explicitly encoding spatiotemporal evolution from historical weather states. Second, VA-MFM adaptively selects frequency thresholds according to the spectral energy distribution of each variable, thereby imposing regularization of comparable strength across all variables. Their joint integration enables the effective application of LDMs to high-resolution ensemble weather forecasting. \ourmodel achieves superior performance to ENS at short lead times while
remaining comparable to ENS at longer horizons. We believe this work establishes a principled foundation for high-resolution, LDM-based forecasting in atmospheric sciences. We further plan to extend this paradigm to high-resolution sub-seasonal ensemble forecasting, investigating the scalability of latent diffusion models under longer lead times and more challenging predictability regimes.

\bibliography{puyun_ldm}
\bibliographystyle{icml2026}




\end{document}